%% file: iclr2026_conference.tex
\documentclass{article}
\usepackage{iclr2026_conference,times}
\input{math_commands.tex}

\usepackage{hyperref}
\usepackage{url}
\usepackage{graphicx}
\usepackage{mathtools}
\usepackage{amsthm}

\usepackage{subcaption}
\usepackage{booktabs}
\usepackage{multirow}

\title{Relative-Based Scaling Law for Neural Language Models}

\author{Baoqing Yue, Jinyuan Zhou, Zixi Wei, Jingtao Zhan\thanks{Corresponding author} , Qingyao Ai, Yiqun Liu \\
Tsinghua University\\
\texttt{\{ybq22,zhoujiny21\}@mails.tsinghua.edu.cn, zixiwei@gmail.com, }\\
\texttt{\{jingtaozhan, aiqy, yiqunliu\}@tsinghua.edu.cn}
}

\iclrfinalcopy
\usepackage{csquotes}
\begin{document}

\maketitle

\begin{abstract}

Scaling laws aim to accurately predict model performance across different scales. Existing scaling-law studies almost exclusively rely on cross-entropy as the evaluation metric. However, cross-entropy provides only a partial view of performance: it measures the absolute probability assigned to the correct token, but ignores the relative ordering between correct and incorrect tokens. Yet, relative ordering is crucial for language models, such as in greedy-sampling scenario.
To address this limitation, we investigate scaling from the perspective of relative ordering. We first propose the Relative-Based Probability (RBP) metric, which quantifies the probability that the correct token is ranked among the top predictions. Building on this metric, we establish the Relative-Based Scaling Law, which characterizes how RBP improves with increasing model size. Through extensive experiments on four datasets and four model families spanning five orders of magnitude, we demonstrate the robustness and accuracy of this law.
Finally, we illustrate the broad application of this law with two examples, namely providing a deeper explanation of emergence phenomena and facilitating finding fundamental theories of scaling laws.
In summary, the Relative-Based Scaling Law complements the cross-entropy perspective and contributes to a more complete understanding of scaling large language models. Thus, it offers valuable insights for both practical development and theoretical exploration.

\end{abstract}

\section{Introduction}

Scaling laws are an important tool in the era of large language models. Their primary goal is to predict how model performance changes as the model size increases. \citep{hestness2017deep,kaplan2020scaling,rosenfeld2019constructive,henighan2020scaling} The key challenge of scaling law studies is identifying reliable performance metrics that can be accurately predicted. \citep{hoffmann2022training} To date, cross-entropy has proven to be the most reliable metric for this purpose, and consequently it has become the dominant choice in scaling law research. Cross-entropy-based scaling laws  not only guide the training of large language models, but also provide insights about model mechanisms and artificial intelligence theories. \citep{kaplan2020scaling,hoffmann2022training,henighan2020scaling} Cross-entropy has even been adopted beyond language modeling to search for scaling laws in new domains, such as multimodal learning and information retrieval. \citep{hackenburg2024evidence,aghajanyan2023scaling,edwards2024scaling,fang2024scaling,shukor2025scaling}

\begin{figure}[t]
    \centering
    \includegraphics[width=1.0\textwidth]{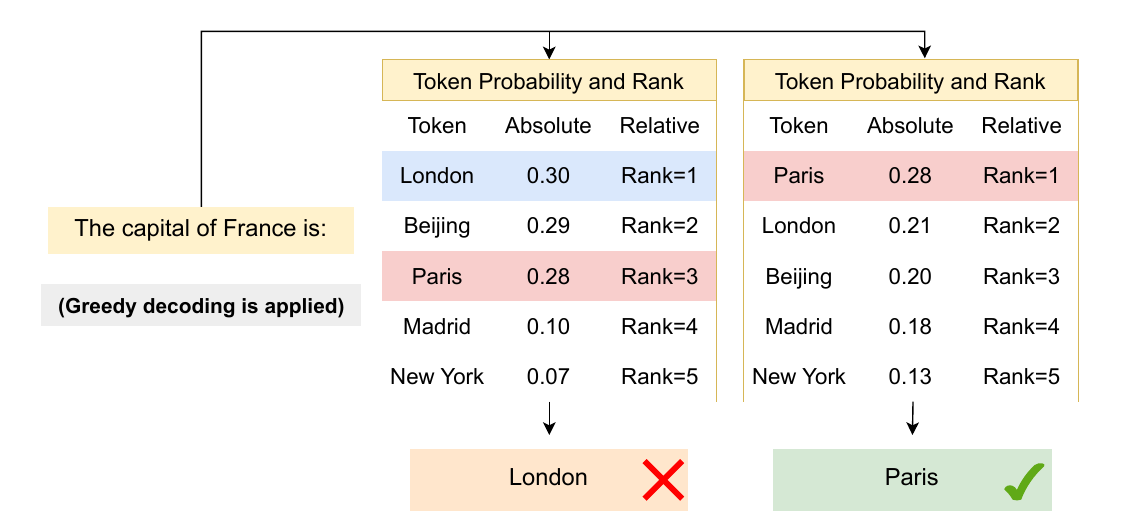} 
    \caption{
        Illustration of the limitation of absolute-based metrics in evaluating generation performance. 
        \textbf{(Left)} The ground-truth token has a high absolute probability but is ranked below competitors, 
        leading greedy decoding to fail. 
        \textbf{(Right)} With the same absolute probability yet a higher rank, the ground-truth token is correctly chosen. 
        This shows that absolute-based metrics cannot capture the crucial relative ranking among tokens.
    }
    \label{fig:intro}
\end{figure}

However, focusing solely on cross-entropy as the metric provides an incomplete picture of a model’s scaling behavior. This limitation arises because cross-entropy primarily measures the absolute probability assigned to the correct answer, while ignoring the relative ordering of predictions. \citep{xu2024understanding,shannon1948mathematical,mackay2003information,zhang2018generalized,wang2019symmetric,ho2019real,feng2021can,mao2023cross,pang2019rethinking,bruch2021alternative,li2021mixed,shim2024enhancing,mezentsev2024scalable} In fact, absolute-based and relative-based perspectives capture two distinct aspects of model performance, and neither can substitute for the other. As illustrated in Figure~\ref{fig:intro}, a model assigns a probability of 0.28 to the correct token. Yet the rank of this token can be different. In one case, two incorrect candidates may still outrank the correct one; in another, all incorrect candidates may receive lower scores, placing the correct token at the top. Thus, the same probability score may correspond to different rankings. Consequently, cross-entropy–based scaling laws fail to capture how the relative position of the correct answer changes with model size. This shortcoming is particularly severe because relative ordering plays a central role in practical applications of language models, such as greedy decoding and top-k sampling. \citep{noarov2025foundations,wei2024diff,bruch2021alternative,tang2024top,ma2025estimating,chatzi2024prediction,freitag2017beam,shi2024thorough,song2407good,prabhu2024pedal,chen2025iterative,naseh2023stealing} The inability to account for such order-sensitive sampling strategies highlights a significant gap between cross-entropy metric and the performance observed in real-world usage.

To address this limitation\footnote{Code is available at \url{https://github.com/ybq22/relative-based-scaling-law}.}, we introduce a new metric, Relative-Based Probability (RBP), to capture a model’s ability to rank the correct token among the top candidates. Given a parameter $k$, RBP\textsubscript{$k$} measures the probability that the correct token appears within model's top-$k$ predictions. The computation of RBP\textsubscript{$k$} proceeds as follows: First, we select a corpus. Then, for each target token in the corpus, we check whether it is included among the top-$k$ tokens predicted by the model. Finally, we calculate the proportion of such cases over the entire dataset, which yields the RBP value. Unlike cross-entropy that is concerned with the absolute prediction value, RBP focuses on the relative ordering of predictions. As a result, RBP\textsubscript{$k$} provides a complementary perspective on model performance.

Next, we examine how RBP\textsubscript{$k$} changes as the model size increases. Our results reveal a clear scaling law when $k$ is much smaller than the vocabulary size. Specifically, we find that
\begin{equation}
    - \log \big(\text{RBP}_{k}\big) \ \propto \ S^{-\alpha}, \quad k \ll \text{Vocab Size},
\end{equation}
where $S$ denotes the number of non-embedding model parameters and $\alpha$ is a positive constant. We refer to this relationship as the Relative-Based Scaling Law. To investigate this relationship, we conduct experiments across four datasets. We use $4$ model families, covering a total of $24$ models whose sizes span over $5$ orders of magnitude. The results show that when $k$ lies in the range of roughly $1$ to $100$, the relationship between $-\log(\text{RBP}_{k})$ and model size can be fitted very well by a power-law. The fitted power-law achieves an $R^2$ of approximately $0.99$, which is comparable to the $R^2$ values obtained from cross-entropy–based scaling laws.
Relative-Based Scaling Law deepens our understanding of scaling behavior: enlarging model size fundamentally reshapes the ranking of tokens. As models grow, a greater proportion of correct tokens are placed among the top predictions, and this improvement follows a power-law trend. Such an understanding goes beyond what cross-entropy–based scaling laws alone can capture.

We believe Relative-Based Scaling Law holds important implications for the development of large language models as well as for advancing the theoretical foundations of artificial intelligence. In this work, we illustrate its utility through two concrete examples.
(1) Relative-Based Scaling Law enhances our understanding of emergence phenomena: Prior studies have attempted to explain emergence through cross-entropy–based scaling laws, but such explanations fail to generalize to decoding strategies like greedy sampling or top-$k$ sampling, where relative ordering is crucial. \citep{wei2022emergent,mckenzie2023inverse,schaeffer2023emergent,lu2023emergent,krakauer2025large} By contrast, the Relative-Based Scaling Law naturally accounts for these scenarios and resolves this issue.
(2) Relative-Based Scaling Law opens new directions for theoretical research: Although RBP and cross-entropy capture complementary aspects of model performance, we find that their scaling behaviors are surprisingly similar. They share not only identical mathematical form but also very close fitted exponents. This confusing coincidence suggests the existence of a more fundamental theory of intelligence that can unify both scaling laws. At present, no established theory can fully explain this phenomenon. In this paper, we put forward a conjecture as a step toward such unification.

In summary, Relative-Based Scaling Law provides a relative-ordering perspective on scaling behavior. It complements the cross-entropy view and contributes to a more complete picture of scaling in neural language models. Together, the two laws shall offer both practical guidance for scaling up language models and theoretical insight into the fundamental theories of artificial intelligence.

\section{Related Work}

Scaling laws for neural language models, often based on absolute metrics like cross-entropy loss, provide a framework for understanding performance evolution with model size, dataset size, and compute budget \citep{kaplan2020scaling,henighan2020scaling}. 

Absolute-based metrics, however, do not capture the relative ranking of the ground-truth token among candidates, which can misalign metric improvements with generative performance \citep{ma2025estimating,wei2024diff,tang2024top}. To address this, relative-based metrics have been proposed, improving rank-sensitive accuracy in both small- and large-scale models \citep{petersen2022differentiable}.

Despite these advances, relative-based evaluation has not been systematically studied under scaling laws for neural language models. Our work fills this gap by introducing a relative-based metric and empirically analyzing its scaling behavior across architectures and datasets.

\section{Relative-based Metric}

In this section, we introduce our metric for evaluating model performance. Prior work has commonly relied on an \emph{absolute-based} approach, which measures the probability assigned to the ground-truth token. In contrast, we propose a new \emph{relative-based} metric, $\text{RBP}_k$, which evaluates whether the ground-truth token appears within model’s top-$k$ predictions. This provides a perspective that is complementary to cross-entropy and is more directly aligned with practical sampling strategies.

\subsection{Prior Absolute-Based Metric}
In previous studies, model performance has typically been evaluated using cross-entropy loss ($\mathcal{L}_{\text{CE}}$), which exhibits clear scaling behavior across model sizes. Let $t$ be the ground-truth token, and ${\rm p_A}(t)$ be the probability score output by the model. Cross-entropy equals
\begin{equation}
    \mathcal{L}_{\text{CE}} = \mathbb{E}[- \log {\rm p_A}(t)],
\end{equation}

However, ${\rm p}(t)$ alone does not reflect the \emph{relative ordering} of the ground-truth token among all candidate tokens. As shown in Figure~\ref{fig:intro}, the ground-truth token may be assigned relatively high probability but not ranked as the top-1 candidate. If greedy-decoding or top-k sampling are used during inference, such a probability score cannot reflect models' real-world performance.

\begin{figure}[t]
    \centering
    \begin{subfigure}[t]{0.48\textwidth}
        \centering
        \includegraphics[width=\textwidth]{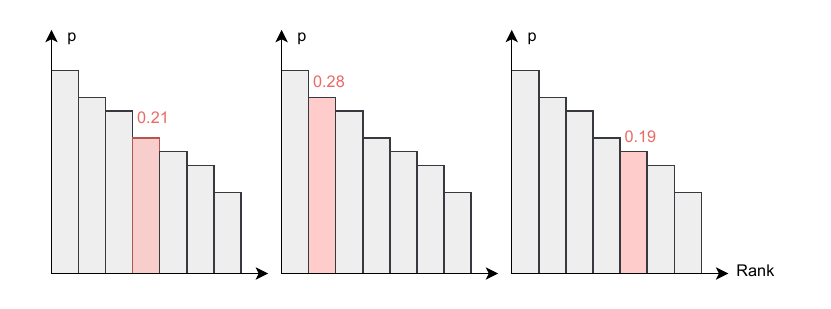}
        \caption{Calculation of the absolute-based metric $\rm p_A$. Ground-truth is highlighted in red. The metric $\rm p_A$ is computed as the average of the probabilities assigned to the ground-truth across different instances, $\rm p_A = (0.21 + 0.28 + 0.19) / 3 \approx 0.227$.}
        \label{fig:absolute_metric}
    \end{subfigure}%
    \hfill
    \begin{subfigure}[t]{0.48\textwidth}
        \centering
        \includegraphics[width=\textwidth]{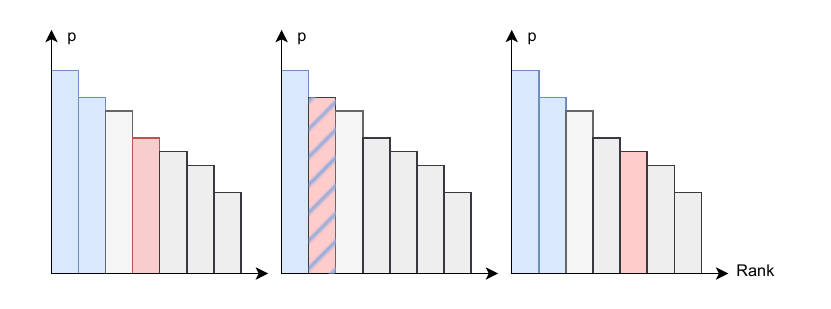}
        \caption{Calculation of the relative-based metric $\text{RBP}_k$ with $k=2$. The ground truth is highlighted in red. The red rectangle with blue hatching indicates cases where the ground truth appears within the top-$2$ ranked predictions. Since the ground truth is included in the top-2 predictions for only one out of the three instances, we obtain $\text{RBP}_2 = 1/3 \approx 0.333$.}
        \label{fig:relative_metric}
    \end{subfigure}
    \caption{Illustration of two distinct metric calculation methods. Figure \ref{fig:absolute_metric} shows the absolute-based metric $\rm p_A$, which directly quantifies the probability assigned to the ground-truth. Figure \ref{fig:relative_metric} shows the relative-based metric $\text{RBP}_{k}$, which measures how often the ground-truth falls within the top-$k$ predictions. These figures highlight how the two metrics capture different aspects of model behavior.}
    \label{fig:combined_metrics}
\end{figure}

\subsection{Our Relative-Based Metric}
To address this problem, we introduce a \emph{relative-based} metric, namely Relative-based Probability.
It is computed based on a hyper-parameter $k$. Therefore, we denote this metric as $\text{RBP}_{k}$. $\text{RBP}_{k}$ measures the probability that the ground-truth token appears within the model’s top-$k$ predictions. Compared to cross-entropy, it is an independent metric that complements the relative-ordering perspective to model performance. For example, if  $\text{RBP}_1$ is $30\%$, it means that greedy sampling works for $30\%$ cases. Yet this cannot be deduced from the cross-entropy value.

$\text{RBP}_{k}$ is formally defined as follows. Let $\mathcal{V}$ denote the vocabulary and $t$ be the ground-truth token. We use ${\rm p}(v)$ be the score of token $v$. Then, the rank of ground-truth token $R$ equals:
\begin{equation}
    R = \sum_{v \in \mathcal{V}} \mathbf{1}\{\rm p(v) \geq p(t)\}.
\end{equation}
$R$ is a random variable. We define $\text{RBP}_{k}$ as its cumulative distribution function. Formally, it equals:
\begin{equation}
    \text{RBP}_{k} = \Pr(R \leq k),
\end{equation}
We can see that $\text{RBP}_{k}$ is the probability that the ground-truth token is within the top-$k$ scored tokens. 

In practice, we compute $\text{RBP}_{k}$ as follows. First, we select a corpus. Then, for each token in the corpus, we record $1$ if the ground-truth token falls in the top-$k$ predictions and $0$ otherwise. The average of these values yields an empirical estimation of $\text{RBP}_{k}$. 

In the next section, we will show that $\text{RBP}_{k}$ also exhibits a power-law scaling behavior with model size when $k$ is small. Therefore it is able to precisely capture performance improvements from a relative-ordering perspective.

\section{Relative-Based Scaling Law}

Based on $\text{RBP}_{k}$, this section establishes the \emph{Relative-based Scaling Law}. It characterizes how $\text{RBP}_{k}$ evolves with model size. 
Our finding is that for small $k$ values ($k \ll |\mathcal{V}|$, with $|\mathcal{V}|$ denoting the vocabulary size), $\text{RBP}_{k}$ follows a precise power-law relationship with model size:
\begin{equation}\label{eq:rv-scaling-general}
-\log \text{RBP}_k \;\propto\; S^{-\alpha} \quad \big( k \ll |\mathcal{V}|\big)
\end{equation}
where $S$ denotes the model size and $\alpha > 0$ is the scaling exponent. We refer to this relationship as the \emph{relative-based scaling law}.  

Now we start to investigate the relationship between $\text{RBP}_{k}$ and model size. Since $\text{RBP}_{k}$ depends on the choice of threshold $k$, experiments are conducted in the following three regimes:

\begin{enumerate}
    \item $k=1$ regime: This regime is corresponding to the hardest setting where only the top-ranked token is considered. In this case, $\text{RBP}_{k}$ closely links to greedy decoding strategy.
    \item Moderate-$k$ regime: In this regime, $k$ is larger than $1$ but far smaller than the vocabulary size. In our experiments, we set $1 < k \leq 100$. This setup closely links to top-k sampling strategies.  
    \item Large-$k$ regime: In this regime, $k$ approaches to the vocabulary size. In our experiments, we set $k$ to $20{,}000$ or $30{,}000$.
 
\end{enumerate}
These three regimes provide a comprehensive view of how $\text{RBP}_{k}$ scales with model size. We present the experimental results respectively in the following three subsections.

\subsection{$k=1$ Regime}  

The result marks the discovery of a novel scaling law. It means that scaling up models increases the probability that the correct token is ranked highest. And the increasing rate follows a power-law rate. This perspective cannot be deduced from prior cross-entropy scaling law. It demonstrates that scaling up models fundamentally changes the relative-ordering and results in more correct tokens ranked at the top. Since $\text{RBP}_1$ closely links to the greedy decoding strategies, it reflects that greedy decoding performance will improve if model is scaled up.

\begin{figure}[t]
    \centering
    \includegraphics[width=1.0\textwidth]{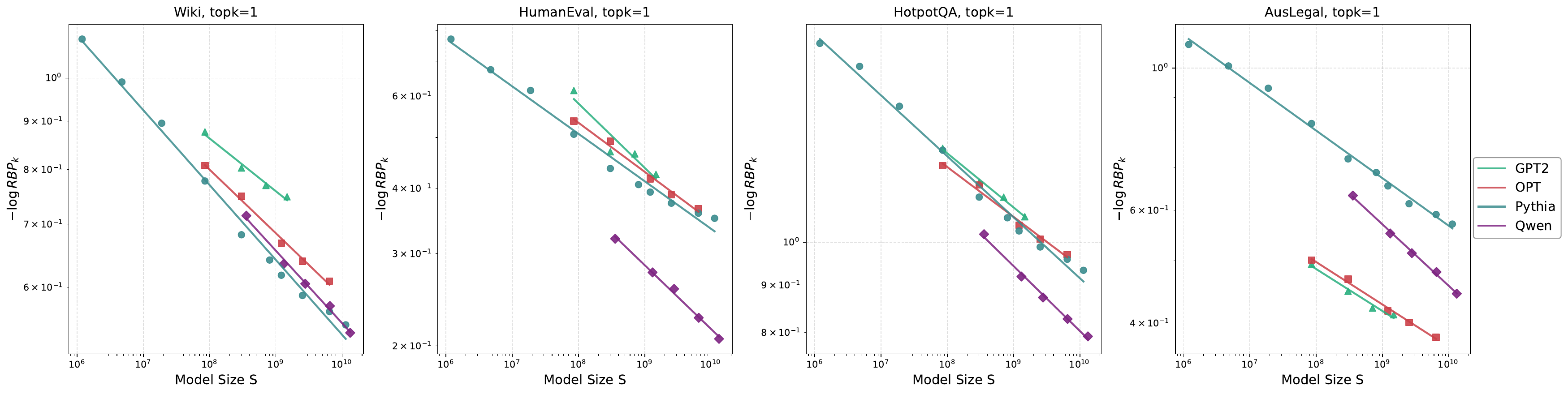}
    \caption{Relative-based scaling laws when $k=1$. Across all model series and all datasets, $\text{RBP}_1$ exhibits precise power-law relationship with model sizes.}
    \label{fig:topk_scaling_1}
\end{figure}

\subsection{Moderate-$k$ Regime: $1 < k \leq 100$} 

\begin{figure}[t]
    \centering
    \includegraphics[width=1.0\textwidth]{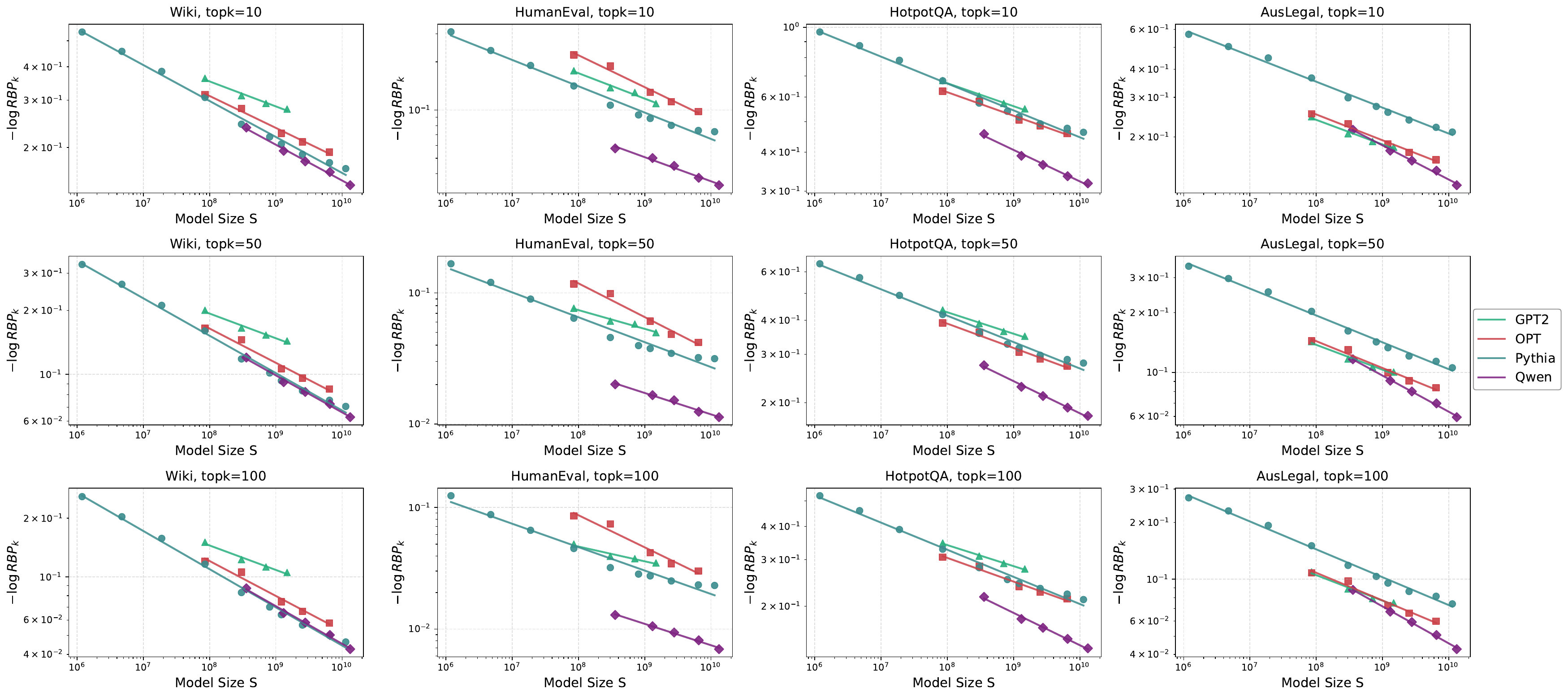}
    \caption{Scaling laws when $1 < k \ll |\mathcal{V}|$. The relative-based metric $\text{RBP}_{k}$ maintains strong power-law scaling behavior, with consistently high $R^2$ values across datasets and model series. This indicates that the metric reliably captures performance improvements under top-$k$ sampling strategies commonly used in practice.}
    \label{fig:topk_scaling_mid}
\end{figure}

Moreover, we also observe that the power-law relationship still holds for small $k$ values.
We set $k$ to $10$, $50$, and $100$. The results are illustrated in Figure~\ref{fig:topk_scaling_mid}. We can see that for all cases, the data points lie on a straight line in the log-log plot. The fitted curves achieve $R^2 \geq 0.97$ across all datasets, indicating that the power-law relationship remains highly robust in this regime. 

The result means that Relative-based Scaling Law can generalize to small $k$ values. When model is scaled up, more and more correct tokens are ranked within the top-$k$ predictions. And the rate of improvement follows a power-law manner. 
Since $\text{RBP}_{k}$ measures the probability that the ground-truth token appears within the top-$k$ candidates, it closely links to top-$k$ sampling strategy. It indicates that top-k sampling result will be improved when model is scaled up, which is another perspective that cannot be deduced from cross-entropy scaling law alone.

\subsection{Large-$k$ Regime: $k \rightarrow |\mathcal{V}|$}  

\begin{figure}[t]
    \centering
    \includegraphics[width=0.9\textwidth]{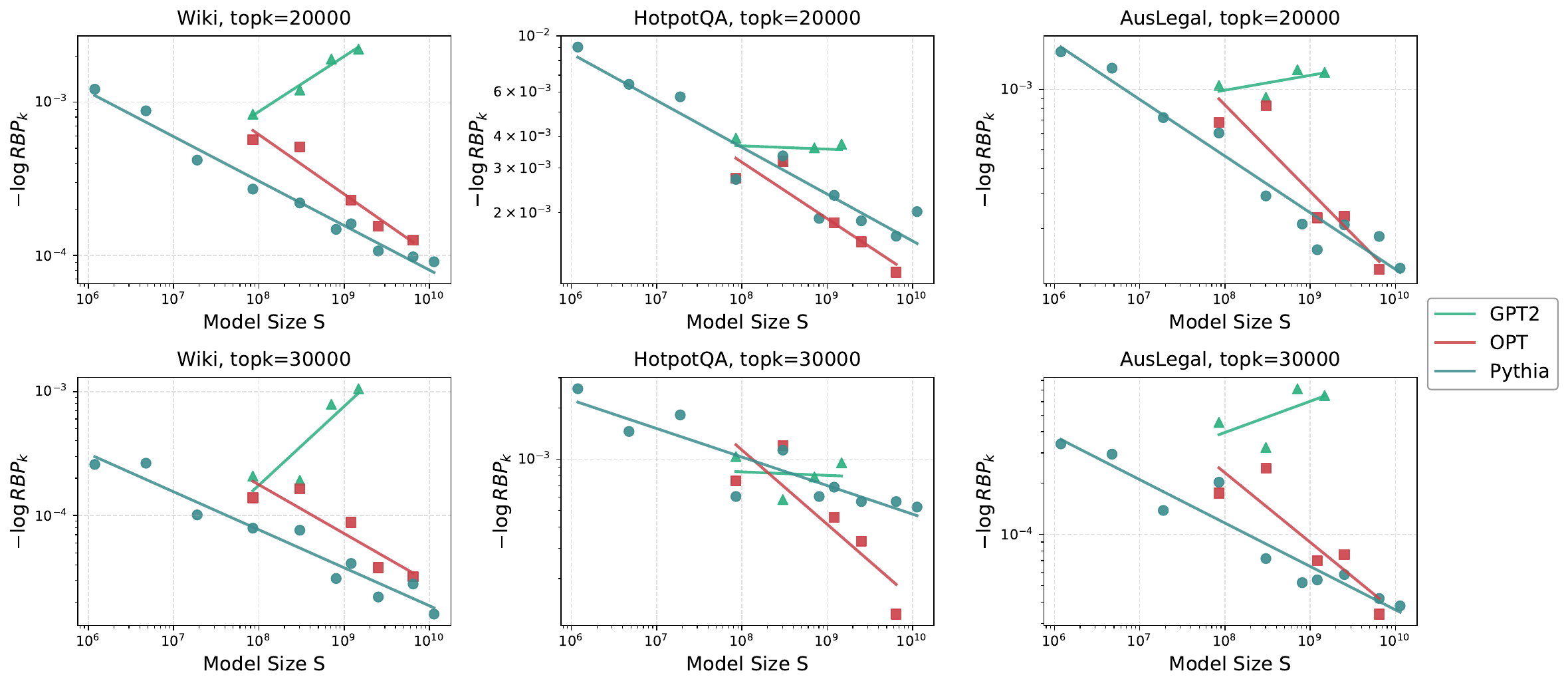}
    \caption{Scaling laws when $k \rightarrow |\mathcal{V}|$. In this regime, the power-law behavior deteriorates substantially, with large scatter and inconsistent slopes across models and datasets. This indicates that $\text{RBP}_{k}$ becomes less informative about scaling when the threshold approaches the vocabulary size.}
    \label{fig:topk_scaling_largek}
\end{figure}

As $k$ approaches the vocabulary size $|\mathcal{V}|$, the scaling behavior of $\text{RBP}_{k}$ breaks down. 
We set $k$ to $20,000$ and $30,000$, and illustrate the results in Figure~\ref{fig:topk_scaling_largek}\footnote{For Pythia, OPT, and GPT-2 models, vocabulary sizes are 50,256, 50,272, and 50,257, respectively, such that the chosen $k$ values nearly cover the entire vocabulary; for Qwen models, vocabulary sizes are 151,936 for models up to 7B and 152,064 for models above 7B. We don't report Qwen series in this regime because it doesn't apply to the condition $k \rightarrow |\mathcal{V}|$.}. We can see that the data points scatter significantly. For example, $\text{RBP}_{k}$ for GPT2 even increases rather than decreases. The case for $k=30,000$ becomes much more random than the case for $k=20,000$. Therefore, we believe that Relative-based Scaling Law only holds for $k$ that is small.

The reason for this behaviour remains to be investigated. 
We conjecture that that reason might lies in random noise. When $k$ nearly covers the entire vocabulary, most ground-truth tokens are included in the top-$k$ set, regardless of model size. In this case, some noisy data may dominate $\text{RBP}_{k}$, which breaks down its power-law scaling pattern. The noise even makes $\text{RBP}_{k}$ increase when model is scaled up, as shown by the GPT2 results in Figure~\ref{fig:topk_scaling_largek}. 

\subsection{Summary}  
\begin{figure}[h!]
    \centering
    \begin{subfigure}[t]{0.48\textwidth}
        \centering
        \includegraphics[width=\linewidth]{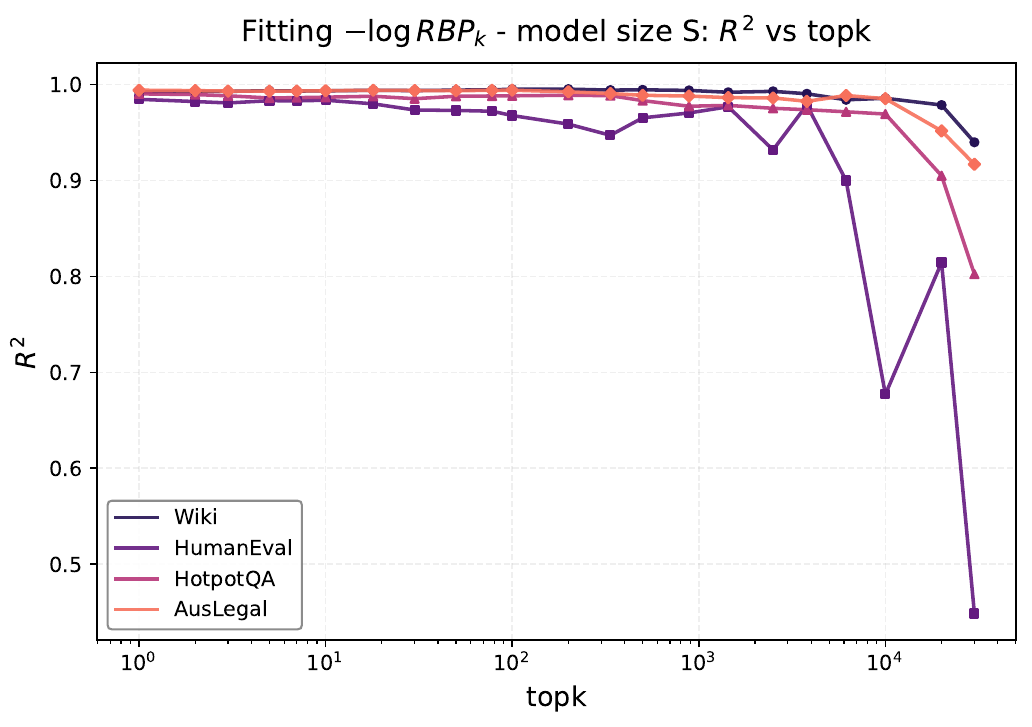}
        \caption{Coefficient of determination ($R^2$) as a function of $k$. The scaling law holds robustly for $k < 1000$, where $R^2$ is very close to $1$. It starts to drop when $k$ is too large.}
        \label{fig:r2-vs-topk}
    \end{subfigure}
    \hfill
    \begin{subfigure}[t]{0.48\textwidth}
        \centering
        \includegraphics[width=\linewidth]{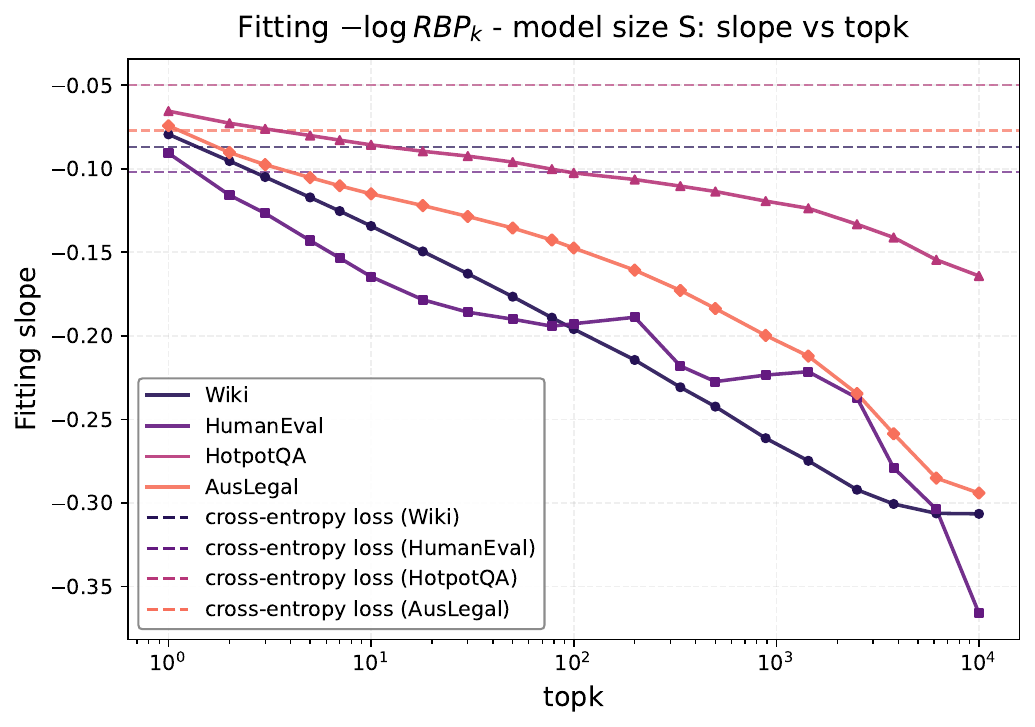}
        \caption{Exponent of the fitted power-law as a function of $k$. Exponent for $k=1$ is close to those of cross-entropy-based scaling laws. The exponent value decreases gradually as $k$ increases, indicating that $\text{RBP}_k$ is easier to optimize for bigger $k$.}

        \label{fig:slope-vs-topk}
    \end{subfigure}
    \caption{Scaling fitting results of $\text{RBP}_{k}$ as a function of $k$. Figure~\ref{fig:r2-vs-topk} shows the power-law fitting quality. Figure~\ref{fig:slope-vs-topk} shows how scaling exponent evolves with $k$.}
    \label{fig:fit-vs-topk}
\end{figure}

The results across the three regimes collectively reveal how the choice of threshold $k$ governs the extent to which $\text{RBP}_{k}$ reflects power-law scaling with model size. In Figure~\ref{fig:fit-vs-topk}, we summarize how the threshold $k$ affects the fitting quality and the scaling exponent.

According to the left figure, we can see that the scaling law holds robustly in the small and moderate $k$ regimes, where $k \ll |\mathcal{V}|$. The fitted quality only starts to drop after $k$ grows close to the vocabulary size. We can see that $R^2$ values remain high (above 0.9) when $k < 1000$. Therefore, Relative-based Scaling Law is a robust law for a wide range of $k$ values.

According to the right figure, the scaling exponent increases as $k$ increases. A higher exponent means that the performance increases faster when the model is scaled up. Therefore, the result indicates that optimizing $\text{RBP}_{100}$ shall be easy, since the exponent is close to $0.2$. Yet it is hard to optimize $\text{RBP}_1$, as the exponent is less than $0.1$. Such an understanding is a unique contribution of Relative-based Scaling Law.

\section{Application of Relative-based Scaling Law}

In this section, we demonstrate two applications of Relative-based Scaling Law. The first is to explain emergence phenomena, and the other is to explore fundamental principles that drive both prior cross-entropy-based and our Relative-based Scaling Laws.

\subsection{Explaining Emergence}
A significant challenge for scaling laws is to explain complex macroscopic phenomena, such as the ``emergence'' of new capabilities in large language models. Emergence refers to the sharp, non-linear jump in performance on a specific task once model size surpasses a threshold. Although \citep{schaeffer2023emergent} explain this phenomenon with cross-entropy-based scaling law, this scaling law cannot generalize to greedy or top-k sampling and thus cannot explain emergence phenomenon in these scenarios. We show that the relative-based scaling law can complement this drawback and provide a quantitative explanation.

\begin{figure}[t]
\centering
\includegraphics[width=\textwidth]{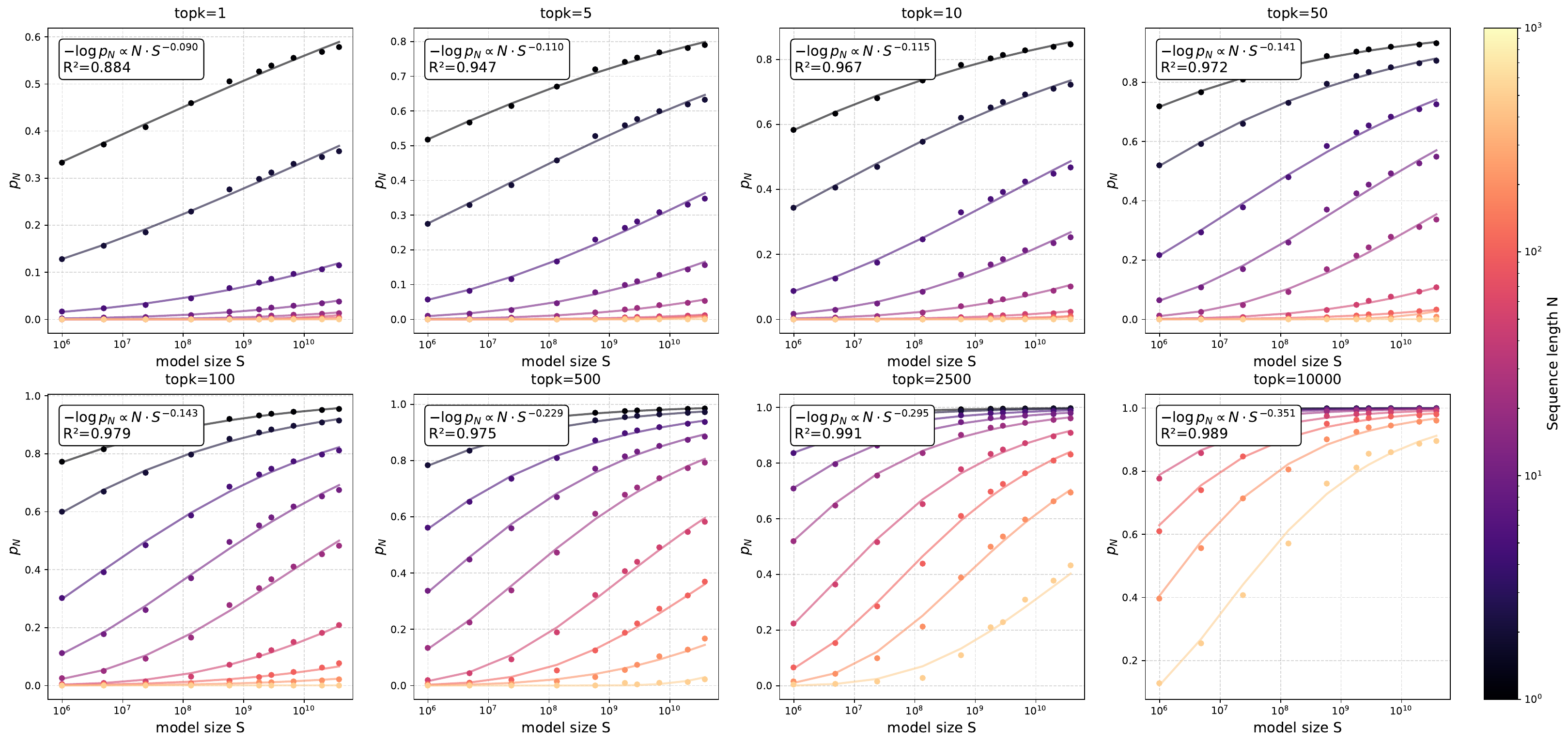}
\caption{
Illustration of the emergence phenomena, $-\log \rm p_{N,k} \propto N \cdot S^{-\alpha}$. Each subplot corresponds to a fixed $k$, with colored lines for different $N$. The fitted power-law curves align closely with empirical data, confirming the law across a wide range of conditions. Crucially, the resulting curves naturally exhibit the sigmoidal shape characteristic of observed emergence phenomena, demonstrating that emergence follows directly from smooth microscopic scaling.
}
\label{fig:scaling_law}
\end{figure}

We define task success as the model correctly predicting $N$ consecutive ground-truth tokens within its top-$k$ candidates. Assuming independence and stationarity across positions, the success probability is:
\begin{equation}
    \rm p_{N,k} = (\text{RBP}_k)^N.
\end{equation}
When $\text{RBP}_k \propto S^{-\alpha}$, the sequence-level relation becomes:
\begin{equation}
    -\log \rm p_{N,k} = -N \log(\text{RBP}_k) \propto N \cdot S^{-\alpha}.
\end{equation}
Thus, sequence-level log-error scales smoothly with model size $S$, linearly amplified by $N$. Figure \ref{fig:scaling_law} confirms this prediction empirically.

This explains the apparent "emergence." Since
\begin{equation}
    \rm p_{N,k} \approx \exp(-C \cdot N \cdot S^{-\alpha}),
\end{equation}
for constant $C$, the probability curve is inherently sigmoidal:
\begin{itemize}
    \item For small $S$, the exponent is large and negative, driving $\rm p_{N,k}$ close to zero.
    \item As $S$ grows, the exponent approaches zero, and $\rm p_{N,k}$ rises sharply toward one.
\end{itemize}
The sequence length $N$ acts as an amplifier: larger $N$ sharpens the transition, creating the observed sudden “knee” in performance. 

In summary, emergence is not a breakdown of scaling laws but a predictable macroscopic effect of a smooth power law at the token level. The perceived phase transition is simply the exponential amplification of microscopic trends when mapped to sequence-level tasks under top-$k$ sampling.

\subsection{Connecting Cross-Entropy- and Relative-Based Scaling Laws}
\label{5.2}
\begin{figure*}[t]
    \centering
    \subfloat[Empirical results: CE and $-\log \text{RBP}_k$ exhibit similar scaling forms and slopes.]{%
        \includegraphics[width=0.45\textwidth]{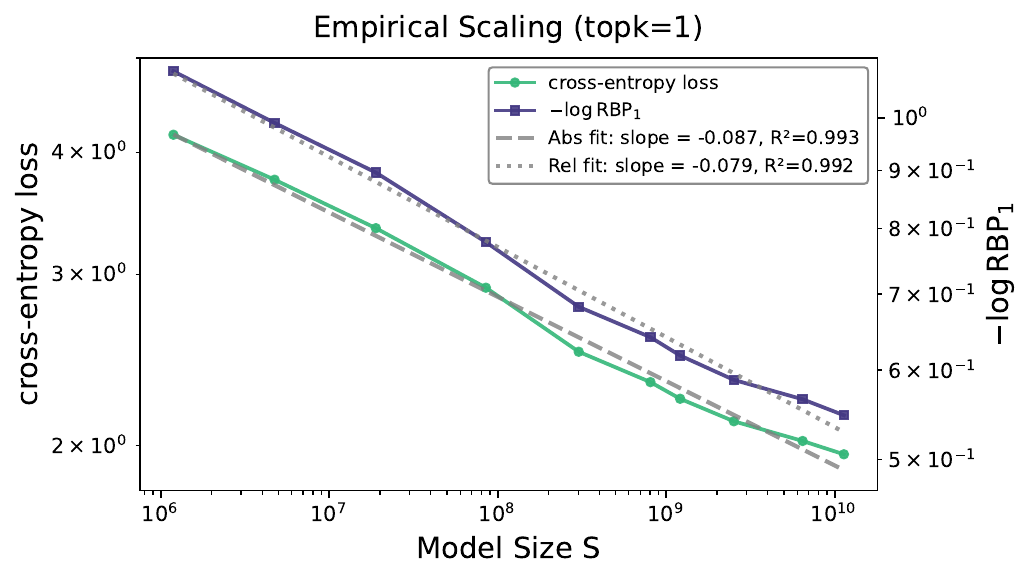}%
        \label{fig:ce_vs_rankingloss}
    }
    \hfill
    \subfloat[Under our lognormal assumption, both the predicted CE and $-\log(\mathrm{RBP}_1)$ exhibit power-law decay with respect to model size, with nearly identical slopes ($|\text{slope}_{\text{CE}} - \text{slope}_{\text{RBP}_1}| < 0.02$, $R^2 > 0.99$)]{%
        \includegraphics[width=0.45\textwidth]{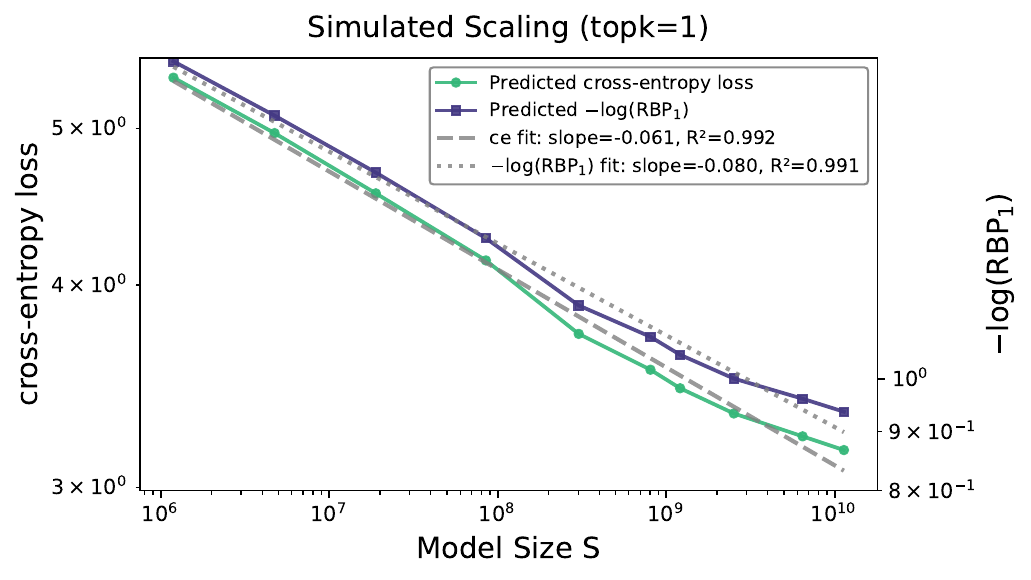}%
        \label{fig:ce_vs_logp1}
    } \\[1em]
    
    \subfloat[Hypothesis: Ground-truth ranking frequency follows the lognormal distribution.]{%
        \includegraphics[width=1\textwidth]{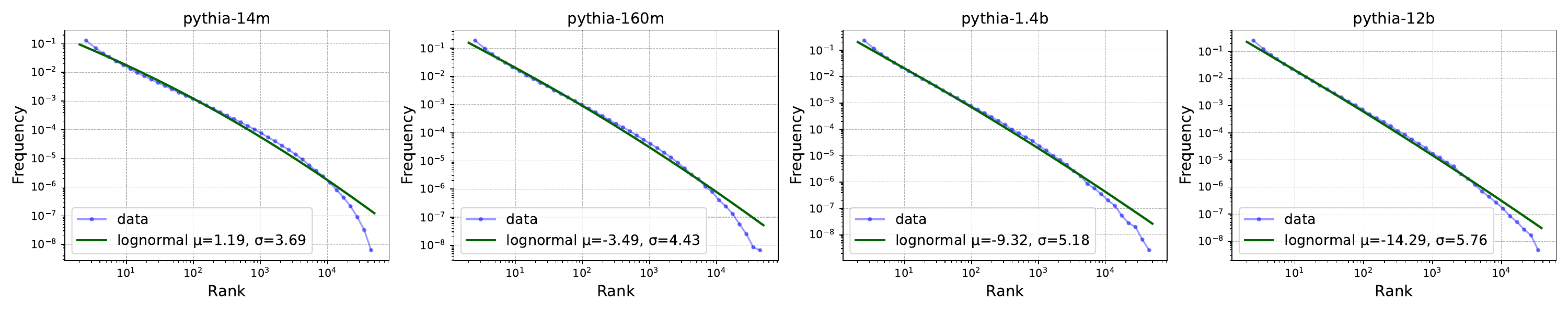}%
        \label{fig:pythia_all_powerlaw}
    }
\end{figure*}

The next application is the peculiar connection between the relative-based scaling law and cross-entropy-based scaling law. As emphasized earlier, the two laws capture fundamentally different aspects of the output distribution: the cross-entropy-based law focuses on the probability mass assigned to the ground-truth token, while the relative-based law examines its rank among candidates. By construction, they are not interchangeable. Nevertheless, we observe that both exhibit remarkably similar power-law decay with respect to model size. In particular, when $k=1$, the decay exponents of cross-entropy (absolute-based) and $-\log p_{1}$ (relative-based) are nearly identical, as shown in Figure~\ref{fig:ce_vs_rankingloss}.

We believe this peculiar coincidence suggests a deeper underlying principle that can derive both laws simultaneously. Yet this perspective has not been proposed in prior theoretical studies, and none of the existing theories can explain this phenomenon. Therefore, we conjecture that this coincidence may hold the key to a unified scaling theory.

To explore this possibility, we analyze the distribution of the ground-truth token's rank across models of different sizes. Empirical evidence shows that these rank distributions are long-tailed~\citep{zhan2025evaluating}. Motivated by this, we assume that the rank distribution follows a lognormal form:
\begin{equation}
    P(x) = \frac{1}{x \sigma \sqrt{2\pi}} 
    \exp\!\left( -\frac{(\ln x - \mu)^2}{2\sigma^2} \right), \quad x > 0,
\end{equation}
where $x$ denotes the rank of the ground-truth token, and $\mu,\sigma$ are parameters that systematically depend on model size $S$. As shown in Figure~\ref{fig:pythia_all_powerlaw}, this assumption is well supported by real data.

Under this assumption, we plot the predicted scaling curves shown in Figure~\ref{fig:ce_vs_logp1}. We treat the parameters $\mu$ and $\sigma$ of the lognormal rank distribution as functions of model size $S$, estimated from real data (Pythia 14M–12B series). For each model, we compute the expected rank distribution of the ground-truth token, assume its probability mass scales by lognormal distribution, and analytically derive both the expected $-\log \text{RBP}_1$ and the corresponding cross-entropy loss (details in Appendix~\ref{appendix:ce_vs_rbp}). Repeating this process across model sizes yields the predicted scaling trends of the two metrics. As shown in Figure~\ref{fig:ce_vs_logp1}, both predicted curves exhibit nearly identical power-law decay with model size, indicating that the cross-entropy and relative-based scaling laws naturally arise from the same underlying lognormal rank mechanism.

This conjecture offers a unified probabilistic framework that connects two previously separate scaling phenomena. By linking the absolute-based and relative-based perspectives through a shared latent rank distribution, it shows that both scaling laws can emerge from a common generative mechanism of model uncertainty. This unified perspective enriches the theoretical understanding of scaling behavior and underscores the value of relative-based metrics as complementary tools for analyzing model performance beyond cross-entropy.

\section{Conclusion}
In this paper, we introduce Relative-based Probability (RBP) and formulate the Relative-Based Scaling Law to analyze model performance from the perspective of relative ordering. Unlike cross-entropy, RBP quantifies the likelihood that the correct token ranks among top predictions, serving as a natural and practically relevant metric for tasks like greedy decoding and top-k sampling.
Experiments across diverse datasets and model families show that this law consistently characterizes performance gains with scale, complementing cross-entropy scaling laws and offering new insights into emergence phenomena and theoretical understanding of scaling behavior.

\bibliography{iclr2026_conference}
\bibliographystyle{iclr2026_conference}

\clearpage
\appendix

\section{Experimental Setup}
\label{appendix:setup}
\paragraph{Models.} We select several representative model families covering a wide range of scales. Specifically, we include the \texttt{Pythia} series \citep{biderman2023pythia} (14M–12B), the \texttt{GPT-2} series \citep{radford2019language}, the \texttt{OPT} family released by Meta \citep{zhang2022opt}, and the more recent \texttt{Qwen2.5} series \citep{qwen2,qwen2.5} (0.5B–14B). Together these span over four orders of magnitude in parameter count and provide multiple independently trained scaling series.

\paragraph{Datasets.} To ensure robustness across domains, we evaluate on four representative benchmarks spanning diverse linguistic and reasoning tasks: \texttt{Wikipedia} for open-domain natural text and general language modeling ability \citep{wikidump}, \texttt{HumanEval} for Python programming tasks focusing on code generation \citep{chen2021evaluating}, \texttt{HotpotQA} for multi-hop question answering that requires handling complex queries \citep{yang2018hotpotqa}, and the \texttt{Open Australian Legal Corpus} for long-form legal documents that test domain-specific expertise \citep{butler-2025-open-australian-legal-corpus}.

\paragraph{Thresholds.} We estimate \(\text{RBP}_k\) across a wide spectrum of threshold values \(k\), spanning over five orders of magnitude, from very small thresholds up to values approaching \(|\mathcal{V}|\). Each data point is obtained by averaging over at least \(5 \cdot 10^5\) tokens to ensure statistical stability.

\section{Additional Results on Scaling Behavior}
As shown in Table~\ref{tab:additional_scaling_results}, we present detailed numerical results for the scaling behavior of different model series across multiple datasets, reported in terms of the fitted $R^2$ and slopes as threshold $k$ increases. To further strengthen the generality of our proposed scaling law, we additionally include results on two large-scale pretraining datasets: \texttt{allenai/C4}\citep{raffel2020exploring,allenai_c4} and \texttt{monology/pile-uncopyrighted}\citep{gao2020pile,pile_uncopyrighted}. We observe clear scaling patterns on all these corpora, which highlights the robustness of the scaling phenomenon. Finally, we note that in most cases the fitted $R^2$ and slope for cross-entropy are very close to those obtained for $-\log \text{RBP}_1$, which motivates our discussion in \ref{5.2}.

\begin{table}[t]
\begin{tabular}{ll|rrrr|rrrr}
\toprule

 & & \multicolumn{4}{c|}{R$^2$} & \multicolumn{4}{c}{Slope} \\
Dataset & Top-$k$ & GPT2 & OPT & Pythia & Qwen & GPT2 & OPT & Pythia & Qwen \\
\cmidrule{3-6} \cmidrule{7-10}
\cmidrule(lr){1-10}
\multirow{7}{*}{AusLegal} & CE  & 0.961 & 0.989 & 0.993 & 0.996 & -0.071 & -0.081 & -0.077 & -0.103 \\
& 1 & 0.980 & 0.994 & 0.994 & 0.997 & -0.066 & -0.066 & -0.074 & -0.096 \\
 & 10 & 0.984 & 0.987 & 0.994 & 0.994 & -0.109 & -0.115 & -0.115 & -0.152 \\
 & 100 & 0.979 & 0.976 & 0.994 & 0.997 & -0.135 & -0.146 & -0.147 & -0.196 \\
 & 500 & 0.982 & 0.970 & 0.989 & 0.996 & -0.150 & -0.181 & -0.184 & -0.218 \\
 & 2500 & 0.905 & 0.945 & 0.986 & 0.971 & -0.184 & -0.219 & -0.235 & -0.260 \\
 & 10000 & 0.822 & 0.924 & 0.985 & 0.901 & -0.077 & -0.340 & -0.294 & -0.230 \\
 
 \cmidrule(lr){1-10}
 
\multirow{7}{*}{C4} & CE  & 0.987 & 0.986 & 0.991 & 0.992 & -0.052 & -0.061 & -0.066 & -0.068 \\
& 1 & 0.985 & 0.993 & 0.991 & 0.991 & -0.049 & -0.055 & -0.063 & -0.062 \\
 & 10 & 0.985 & 0.987 & 0.992 & 0.992 & -0.087 & -0.100 & -0.108 & -0.110 \\
 & 100 & 0.985 & 0.977 & 0.992 & 0.993 & -0.121 & -0.148 & -0.157 & -0.163 \\
 & 500 & 0.989 & 0.962 & 0.992 & 0.994 & -0.145 & -0.185 & -0.193 & -0.203 \\
 & 2500 & 0.985 & 0.946 & 0.992 & 0.986 & -0.156 & -0.222 & -0.230 & -0.228 \\
 & 10000 & 0.974 & 0.911 & 0.988 & 0.953 & -0.162 & -0.260 & -0.258 & -0.237 \\

\cmidrule(lr){1-10}
 
\multirow{7}{*}{Github} & CE  & 0.091 & 0.987 & 0.995 & 0.998 & -0.027 & -0.108 & -0.146 & -0.133 \\
& 1 & 0.959 & 0.980 & 0.995 & 0.998 & -0.076 & -0.096 & -0.134 & -0.118 \\
 & 10 & 0.885 & 0.990 & 0.996 & 0.999 & -0.084 & -0.140 & -0.189 & -0.185 \\
 & 100 & 0.770 & 0.975 & 0.995 & 0.998 & -0.072 & -0.175 & -0.239 & -0.242 \\
 & 500 & 0.671 & 0.936 & 0.995 & 0.998 & -0.062 & -0.199 & -0.277 & -0.269 \\
 & 2500 & 0.083 & 0.895 & 0.994 & 0.998 & -0.012 & -0.242 & -0.323 & -0.326 \\
 & 10000 & 0.045 & 0.862 & 0.980 & 0.976 & -0.002 & -0.244 & -0.363 & -0.338 \\

 \cmidrule(lr){1-10}
\multirow{7}{*}{HotpotQA} & CE  & 0.997 & 0.983 & 0.992 & 0.996 & -0.044 & -0.046 & -0.050 & -0.063 \\
& 1 & 0.993 & 0.987 & 0.990 & 0.995 & -0.058 & -0.054 & -0.066 & -0.070 \\
 & 10 & 0.987 & 0.990 & 0.987 & 0.991 & -0.073 & -0.075 & -0.086 & -0.100 \\
 & 100 & 0.995 & 0.981 & 0.988 & 0.994 & -0.080 & -0.089 & -0.103 & -0.122 \\
 & 500 & 0.988 & 0.954 & 0.983 & 0.992 & -0.073 & -0.096 & -0.114 & -0.145 \\
 & 2500 & 0.950 & 0.954 & 0.975 & 0.992 & -0.067 & -0.113 & -0.133 & -0.166 \\
 & 10000 & 0.989 & 0.868 & 0.969 & 0.973 & -0.073 & -0.109 & -0.164 & -0.150 \\

 \cmidrule(lr){1-10}
\multirow{7}{*}{HumanEval} & CE  & 0.905 & 0.984 & 0.983 & 0.988 & -0.121 & -0.127 & -0.102 & -0.120 \\
& 1 & 0.895 & 0.989 & 0.985 & 0.995 & -0.123 & -0.094 & -0.091 & -0.122 \\
 & 10 & 0.979 & 0.984 & 0.984 & 0.984 & -0.159 & -0.199 & -0.165 & -0.150 \\
 & 100 & 0.947 & 0.966 & 0.968 & 0.994 & -0.122 & -0.264 & -0.193 & -0.176 \\
 & 500 & 0.930 & 0.859 & 0.965 & 0.938 & -0.145 & -0.253 & -0.227 & -0.221 \\
 & 2500 & 0.947 & 0.857 & 0.932 & 0.846 & -0.202 & -0.153 & -0.237 & -0.188 \\
 & 10000 & 0.630 & 0.951 & 0.677 & 1.000 & -0.217 & -0.394 & -0.366 & 0.000 \\
 \cmidrule(lr){1-10}
\multirow{7}{*}{Wiki} & CE  & 0.965 & 0.988 & 0.993 & 0.993 & -0.058 & -0.077 & -0.087 & -0.085 \\
& 1 & 0.983 & 0.994 & 0.992 & 0.994 & -0.056 & -0.067 & -0.079 & -0.077 \\
 & 10 & 0.982 & 0.991 & 0.993 & 0.995 & -0.093 & -0.119 & -0.134 & -0.134 \\
 & 100 & 0.979 & 0.981 & 0.995 & 0.995 & -0.126 & -0.179 & -0.196 & -0.193 \\
 & 500 & 0.950 & 0.968 & 0.994 & 0.993 & -0.133 & -0.228 & -0.242 & -0.234 \\
 & 2500 & 0.826 & 0.942 & 0.993 & 0.985 & -0.084 & -0.273 & -0.292 & -0.266 \\
 & 10000 & 0.915 & 0.917 & 0.986 & 0.942 & 0.090 & -0.331 & -0.306 & -0.325 \\
\bottomrule
\end{tabular}
\caption{Fitted $R^2$ and slopes of $- \log \text{RBP}_k$ as a function of model size $S$ for different datasets and top-$k$ thresholds.}
\label{tab:additional_scaling_results}
\end{table}
\clearpage

\section{Connection Between Absolute-Based and Relative-based Scaling Law}
\label{appendix:ce_vs_rbp}

We assume that the token probabilities $\rm p_k$ follow a lognormal distribution:
\begin{equation}
\rm p_k = \frac{\psi(k)}{c(\mu,\sigma)}, \quad \text{where} \quad
\psi(x) = \frac{1}{\sqrt{2\pi}\,\sigma\,x} \exp\Big(-\frac{(\ln x-\mu)^2}{2\sigma^2}\Big), \quad x>0,
\end{equation}
and $c(\mu,\sigma)$ is the normalizing factor:
\begin{equation}
c(\mu,\sigma) = \sum_{k=1}^\infty \psi(k).
\end{equation}

Define the standard normal pdf and cdf as
\begin{equation}
\varphi(t) = \frac{1}{\sqrt{2\pi}} e^{-t^2/2}, \qquad \Phi(t) = \int_{-\infty}^t \varphi(u)\,du,
\end{equation}
and let
\begin{equation}
a = -\frac{\mu}{\sigma}, \qquad P = \int_1^\infty \psi(x)\,dx = \Phi\Big(\frac{\mu}{\sigma}\Big).
\end{equation}
Conveniently, $\psi(1) = \varphi(a)/\sigma$.

We also define the truncated moments (using $t = \ln x$):
\begin{align}
I_1 &\equiv \int_1^\infty \psi(x)\,\ln x\,dx = \mu P + \sigma \varphi(a),\\
I_2 &\equiv \int_1^\infty \psi(x)\,\ln^2 x\,dx = (\mu^2 + \sigma^2) P + \mu \sigma \varphi(a).
\end{align}

Since $\rm p_1 = \psi(1)/c$, with $\psi(1) = \frac{1}{\sqrt{2\pi}\sigma} e^{-\mu^2/(2\sigma^2)}$, we have
\begin{equation}
-\log \rm p_1 = \log c(\mu,\sigma) + \log(\sqrt{2\pi}\sigma) + \frac{\mu^2}{2\sigma^2}.
\end{equation}

Starting from the definition of cross-entropy,
\begin{equation}
\mathrm{CE} = -\sum_{k\ge 1} \rm p_k \log \rm p_k
= \log(\sqrt{2\pi}\sigma) + \log c(\mu,\sigma) + \frac{1}{c(\mu,\sigma)} \Bigg[ \frac{1}{2\sigma^2}\sum_k \psi(k) (\ln k - \mu)^2 + \sum_k \psi(k) \ln k \Bigg],
\end{equation}
and using the integral approximations $\sum_k \psi(k) \ln k \approx I_1$, $\sum_k \psi(k) \ln^2 k \approx I_2$, and $c(\mu,\sigma) \approx \tfrac12 \psi(1) + P$, we obtain
\begin{equation}
\mathrm{CE} \approx \log(\sqrt{2\pi}\sigma) + \log c + \frac{1}{c} \Bigg[ \frac{1}{2\sigma^2} (I_2 - 2\mu I_1 + \mu^2 c) + I_1 \Bigg].
\end{equation}

Substituting $(I_1,I_2)$ gives
\begin{equation}
\mathrm{CE} \approx \log(\sqrt{2\pi}\sigma) + \log c + \frac{\mu^2}{2\sigma^2} + \frac{1}{c} \Bigg[ P\Big(\frac{\mu^2+\sigma^2}{2\sigma^2} + \mu\Big) + \varphi(a)\Big(\frac{\mu}{2\sigma} + \sigma\Big) - \frac{\mu}{\sigma^2}(\mu P + \sigma \varphi(a)) \Bigg],
\end{equation}
with
\begin{equation}
c \approx \tfrac12 \psi(1) + P = \tfrac12 \frac{\varphi(a)}{\sigma} + P.
\end{equation}

Simplifying yields
\begin{equation}
\mathrm{CE} \approx \log(\sqrt{2\pi}\sigma) + \log c + \frac{\mu^2}{2\sigma^2} + \frac{1}{c} \Bigg[ P\Big(\frac{\sigma^2 - \mu^2}{2\sigma^2} + \mu\Big) + \varphi(a)\Big(\sigma - \frac{\mu}{2\sigma}\Big) \Bigg].
\end{equation}

Since 
\begin{equation}
\Phi(x) = o(\varphi(x)) \quad \text{as } x \to -\infty,
\end{equation}
and
\begin{equation}
\psi(1) = \frac{\varphi(a)}{\sigma} = o(\varphi(a)) \quad \text{as } \sigma \to \infty,
\end{equation}
we have
\begin{equation}
c(\mu, \sigma) = o\!\big(\varphi(\tfrac{\mu}{\sigma})\big).
\end{equation}
Thus, as the model scales, eventually
\begin{equation}
-\log c > -\log(\varphi(a)) = \tfrac12 \log(2\pi) + \tfrac{\mu^2}{2\sigma^2},
\end{equation}
So $\log c$ decays  faster than the term $\tfrac{\mu^2}{2\sigma^2}$. In comparison, the factor $\log(\sqrt{2\pi}\sigma)$ increases only slowly with $\sigma$, and moreover $\sigma$ itself grows only mildly with model size. The remaining correction terms in CE, namely the fraction
\begin{equation}
\frac{1}{c}\Bigg[ P\Big(\frac{\sigma^2 - \mu^2}{2\sigma^2} + \mu\Big) + \varphi(a)\Big(\sigma - \frac{\mu}{2\sigma}\Big) \Bigg],
\end{equation}
do not dominate either: the numerator grows at most polynomially in $\mu$ and $\sigma$, while the denominator $c$ shrinks super-exponentially (since $c = o(\varphi(\mu/\sigma))$). Hence the whole fraction is asymptotically negligible compared to $\log(c)$. Therefore, as $|\mu|$ increases, $\log(c)$ becomes the dominant part of CE. Importantly, this also applies to $-\log(\rm p_1)$, which explains why CE and $-\log(\rm p_1)$ behave similarly as the model scales.
\end{document}

%% file: math_commands.tex
\usepackage{amsmath,amsfonts,bm}

\def\eqref#1{equation~\ref{#1}}

\def\1{\bm{1}}

\DeclareMathAlphabet{\mathsfit}{\encodingdefault}{\sfdefault}{m}{sl}
\SetMathAlphabet{\mathsfit}{bold}{\encodingdefault}{\sfdefault}{bx}{n}